\documentclass[pmlr]{jmlr}

\usepackage{longtable}

\usepackage{booktabs}
\usepackage[load-configurations=version-1]{siunitx} 

 \usepackage{comment}
 \usepackage{url}
 \usepackage{caption}
 \usepackage{listings}
\usepackage{multirow}

\newcommand{\ml}[1]{\textcolor{orange}{\bf [lan: #1]}}
\newcommand{\nish}[1]{\textcolor{magenta}{\bf [nish: #1]}}

\theorembodyfont{\upshape}
\theoremheaderfont{\scshape}
\theorempostheader{:}
\theoremsep{\newline}

\jmlrvolume{1}
\jmlryear{2010}
\jmlrworkshop{AI4ED Workshop at AAAI}

\definecolor{codegreen}{rgb}{0,0.6,0}
\definecolor{codegray}{rgb}{0.5,0.5,0.5}
\definecolor{codepurple}{rgb}{0.58,0,0.82}

\title[Test Case Generation]{Using Large Language Models for Student-Code Guided \\ Test Case Generation in Computer Science Education}

    \author{\Name{Nischal Ashok Kumar} \Email{nashokkumar@cs.umass.edu} \\
   \Name{Andrew S Lan} \Email{andrewlan@cs.umass.edu}\\
   \addr University of Massachusetts Amherst}

\begin{document}

\maketitle

\begin{abstract}

In computer science education, test cases are an integral part of programming assignments since they can be used as assessment items to test students' programming knowledge and provide personalized feedback on student-written code. The goal of our work is to propose a fully automated approach for test case generation that can accurately measure student knowledge, which is important for two reasons. First, manually constructing test cases requires expert knowledge and is a labor-intensive process. Second, developing test cases for students, especially those who are novice programmers, is significantly different from those oriented toward professional-level software developers. Therefore, we need an automated process for test case generation to assess student knowledge and provide feedback. In this work, we propose a large language model-based approach to automatically generate test cases and show that they are good measures of student knowledge, using a publicly available dataset that contains student-written Java code. We also discuss future research directions centered on using test cases to help students.

\end{abstract}

\section{Introduction}
\label{sec:intro}

Large language models (LLMs) have shown great promise in advancing the field of education by helping instructors create educational resources like quizzes, and questions for domains like mathematics, programming, and language learning/ comprehension \citep{baidoo2023education,mcnichols2023exploring,sarsa2022automatic,ashok-kumar-etal-2023-improving}. Particularly, in computer science (CS) education, researchers are leveraging LLMs to automatically generate programming exercises, and provide code explanations, and personalized feedback to students. The work in \citep{sarsa2022automatic} uses the OpenAI Codex model\footnote{\url{https://openai.com/blog/openai-codex}} to aid educators in teaching a course, significantly reducing their manual effort. The work in \citep{10.1145/3544548.3580919} has shown that using Codex in instruction can help improve the students' ability to learn programming. Their study showed that Codex significantly increased code-authoring performance and improved student performance on post-tests in the long term. The work in \citep{10.1145/3587102.3588785} uses LLMs to generate code explanations that serve as examples to improve students' ability to understand and explain code. Their study shows that LLM-generated code explanations are much more accurate and significantly easier to understand than student-written explanations. The work in \citep{liffiton2023codehelp} develops an LLM-based system, CodeHelp, that provides on-demand assistance to programming students without directly revealing solutions to them. Their study shows that CodeHelp is received well by both students and educators, helping students resolve their errors better and complementing educators' teaching efforts. 
The work in \citep{10.1145/3568812.3603476} evaluates two LLMs, OpenAI's ChatGPT and GPT-4 on six tasks: program repair, hint generation, grading feedback, peer programming, task synthesis, and contextualized explanation. Their study showed that GPT-4 performs significantly better than ChatGPT and can be as good as human tutors on some tasks. The work in \citep{phung2023automating} designs a human tutor-style programming feedback system that leverages GPT-4 (the ``tutor'') to generate hints using the information of failing test cases and validates the hint quality using a GPT-3.5 model (``the student''). Their evaluation on three diverse Python-based datasets showed that the precision of their system is close to that of human tutors. 

Despite the recent focus on using LLMs in CS education research, relatively few works have focused on generating \emph{test cases}. The work in \citep{agarwal2022legent} proposes a method, LEGenT, for generating test cases as targeted feedback for an introductory programming course. They use a programming language-centric approach for generating targeted test cases and use them as feedback for students. Although their approach generalizes across 11 programming languages, their work has a key limitation: 
they generate test cases at the code level, i.e., test cases that a piece of student-written code will fail on, without considering generating test cases at the problem level, i.e., generating a set of test cases to measure students' programming knowledge from the code they write for the problem. Therefore, their approach is useful for targeted feedback but ignores the value of test cases as assessments. We need a scalable way to automatically generate test cases for (especially novice) students since they exhibit a wide range of errors, which makes manually generating test cases impractical. 


To bridge the gap in the literature on automated test case generation in educational settings, we propose an LLM-based approach for student code-aware test case generation for programming problems. Our major contributions are: 
\begin{itemize}
    \item We propose a fully automatic iterative refinement-based approach for test case generation that leverages representative student codes, using both LLMs and code compilers\footnote{The code for our paper can be found at: \url{https://github.com/umass-ml4ed/test_case_generation}}. 
    \item We evaluate our approach on the publicly available CSEDM Challenge dataset and show that we can generate test cases that accurately measure student knowledge. 
\end{itemize}

\section{Related Work}
\label{sec:realted-work}

Several works in the domain of software engineering explore the generation of test cases using LLMs. \citep{10172800} proposes to use OpenAI Codex for improving search-based-software-testing systems by using Codex to generate test cases for under-covered functions. \citep{vikram2023can} explores the use of LLMs for generating property-based tests for testing code that implements functions of certain Python libraries like NumPy. The motivation and problem setting of these works are not the same as ours and their goal is not to generate test cases to measure student knowledge in educational settings. In the domain of education, \citep{sarsa2022automatic} conducts preliminary experiments using Codex to generate test cases. Their approach of simply prompting the model sometimes generates faulty test cases as unlike our approach they do not have a feedback mechanism to improve the generated test cases. \citep{agarwal2022legent} proposes a programming language-centric method for generating test cases, LEGenT. LEGenT compares buggy student code with correct code to generate test cases that are used to provide personalized feedback to students. Their approach relies heavily on the structural similarities of the buggy code and the correct code and does not work on complex concepts like arrays, pointers, and recursion. Since we use a LLM based system our method is more generic and not heavily dependent on the structure of the code. Besides, their method can be used to only generate test cases that do not pass the buggy code, whereas the goal of our system is to generate test cases that both pass as well as fail a particular code based on some constraints.

\begin{figure}[htbp]
\floatconts
  {fig:approach}
  {\caption{Visualizing our overall approach for test case generation. TC stands for test case.}}
  {\includegraphics[width=\linewidth]{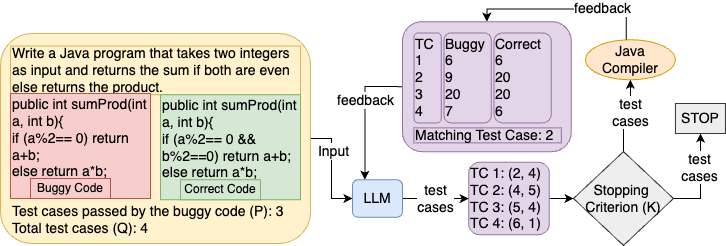}}
\end{figure}

\section{Test Case Generation Methodology}

We now detail the overall prompting process, summarized in \equationref{eq:problem-definition}, for test case generation. Our approach uses iterative refinement: in each iteration, we select a pair of representative student code samples for a given problem to help the LLM generate a set of $Q \leq T$ meaningful and relevant test cases. 

\begin{align}
    \{t\}_{q=1}^T = LLM(prb, \{code\}_{j=1}^C) \label{eq:problem-definition}
\end{align}

\paragraph{Code Pair Prompting}
In each iteration, we design a prompt for the $LLM$ consisting of i) the programming question and ii) a pair of codes $(c_j, c'_j)$, which represent buggy student code and the correct student code, respectively, as shown in \equationref{eq:code-pair}. Let $s_j \in [0,1)$ denote the scores for the buggy code $c_j$; we instruct the $LLM$ to specifically generate test cases $\{t\}_{q=1}^Q$ such that on executing the test cases, the correct code $c'_j$ passes all $Q$ test cases and the buggy code $c_j$ passes exactly $P = Q \times s_i$ test cases. This way, we instruct the LLM to generate test cases that are responsive to common errors contained in actual student code with a certain degree of correctness, which is reflected by the target score $s_j$. 
To improve the test case generation ability of the $LLM$, we prompt the model to generate test cases in a Chain-of-Thought \citep{wei2022chain} manner, by asking it to first output an explanation of the bug in the code and then generate a test case that will not pass the code because of the existence of that bug.

\begin{align}
    \{t\}_{q=1}^Q = LLM(prb, c_j, c'_j)\label{eq:code-pair}
\end{align}

\paragraph{Code Pair Selection}
\label{sec:code-selection}
To capture common errors students make, we devise a strategy to select \emph{three} code pairs $c_j$ and $c'_j$ since using all student codes is impractical. For $c'_j$, we randomly sample a fully-correct student code. For buggy codes, we first sample a code whose score is equal to the median of all student codes. We then sample a code whose score is closest to the full score but not perfect. We then sample a code whose score is the median of the scores between the first two sampled codes. This procedure ensures that we choose representative student codes with few errors, which increases the coverage of the test cases to capture individual bugs since it is difficult to ask the LLM to generate test cases for codes with low scores that may have many bugs or are substantially wrong. 

\paragraph{Compiler Feedback and Iterative Refinement}
Since the $LLM$ by itself cannot execute the generated test cases, 
we use a code compiler to automatically execute both the buggy code $c_j$ and the correct code $c'_j$ against the $Q$ test cases generated by the $LLM$ to obtain the \emph{estimated} scores for each code. Then, we construct a feedback prompt \citep{madaan2023self} with this information and re-prompt the $LLM$ to update the set of test cases (if necessary). This feedback process encourages the LLM to adjust its initial generated test cases using code compiler feedback. We repeat this process for $K$ iterations to get a total of $T \geq Q$ unique test cases.



\section{Experiments}

We now evaluate our approach on a publicly available student code dataset. 

\paragraph{Datasets, Setup, and Metrics} We use CSEDM challenge dataset\footnote{\url{https://sites.google.com/ncsu.edu/csedm-dc-2021/home}}, which contains time-stamped code submissions from over 300 students to 50 Java coding problems grouped by 5 assignments, each of which requires 10-26 lines of code. Each code is scored between 0 and 1, which corresponds to the fraction of test cases passed by the student code.
We use GPT-4 as the underlying LLM since GPT-3.5-turbo and smaller models are not as good at following the instructions in the prompt. We set the temperature of the generation to 0 and the maximum sequence length to 1000 tokens. 
We use a total of $K=1$ iterations since GPT-4 is good at incorporating compiler feedback; more iterations are necessary for smaller models. 
We evaluate whether our approach can accurately score student-written code; $\hat{s}_{i,j}$ denotes the estimated score for code $c_{i,j}$ using the generated test cases $\{t\}^{T}_{q=1}$. Here, $I(c_{i,j},t_q)$ is a binary-valued function which is $1$ if student $j$'s code for problem $i$ passes test case $t_q$ and 0 otherwise. We consider all the code attempts of student $j$ for the same problem $i$ and do not explicitly index attempts for notation simplicity. We report the error $e_{i,j}$ between the estimated score and the ground truth score, as in \equationref{eq:error}. 

\begin{align}
    \hat{s}_{i,j} = \textstyle \sum_{q=1}^T I(c_{i,j},t_q) / T, \quad e_{i,j} = |s_{i,j} - \hat{s}_{i,j}|.\label{eq:error}
\end{align}

\paragraph{Results and Discussion}

\begin{table}[hbtp]
\begin{minipage}{0.45\textwidth}
  \small
  \centering
  \captionsetup{type=figure} 
  \floatconts
    {tab:results}
    {\caption{Mean error across problems for each assignment.}}
    {\scalebox{.9}{\begin{tabular}{lll}
    \toprule
    \bfseries Assignment & \bfseries Data Type & \bfseries Mean \\
    \midrule
    439 & int and boolean & 0.1751\\
    487 & int & 0.2700\\
    492 & string & 0.1289\\
    494 & int array & 0.1475\\
    502 & int array & 0.1008\\ 
    \bottomrule
    \end{tabular}}}
\end{minipage}%
\begin{minipage}{0.55\textwidth}
  \tiny
  \centering
   \captionsetup{type=figure}
  \floatconts
    {fig:494-46}
        {\caption{Error and frequency distribution for student code with different scores for one problem.}}
    {\includegraphics[width=\linewidth]{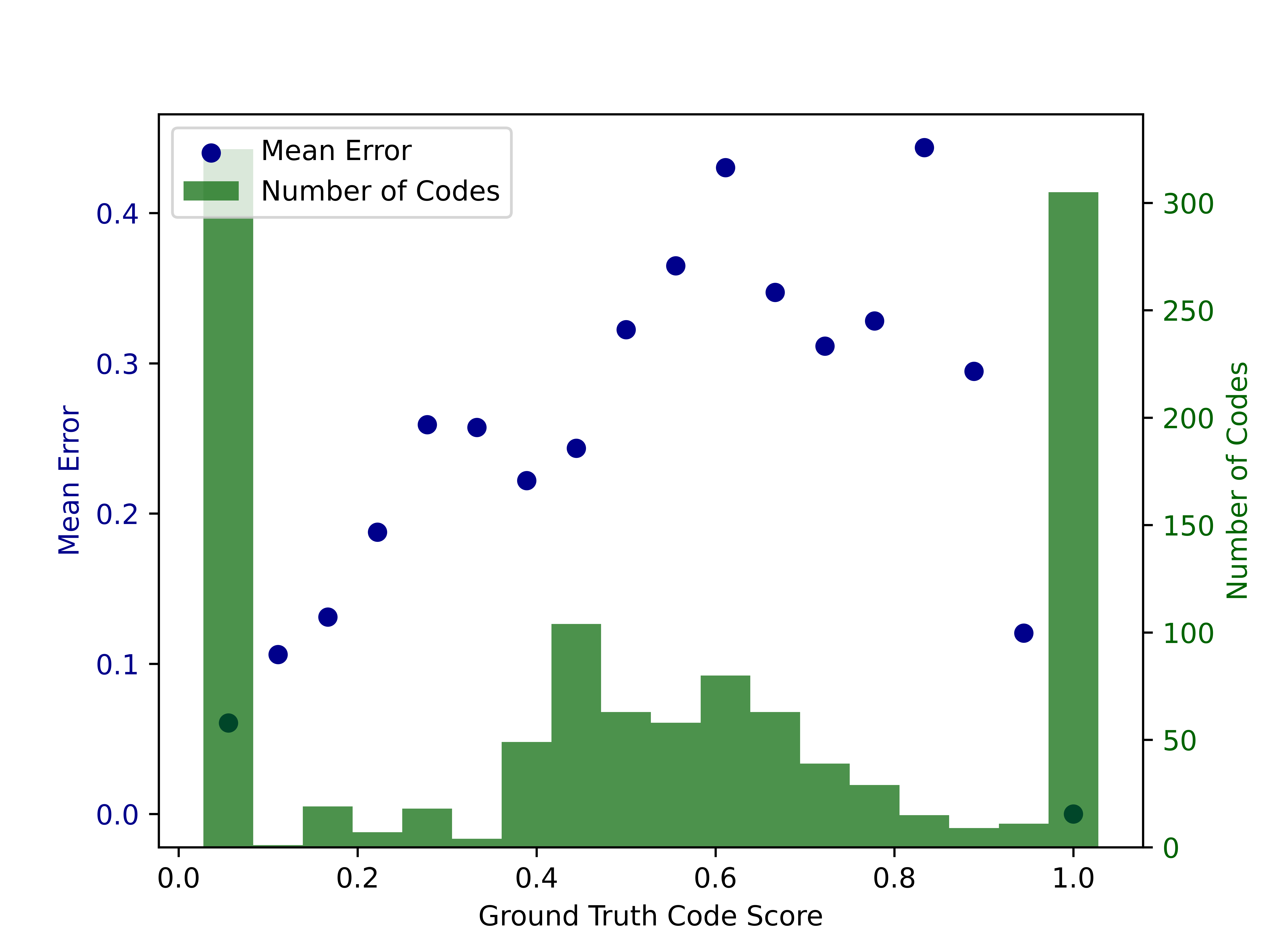}}
\end{minipage}
\end{table}

\tableref{tab:results} shows the input data type and the mean of errors across all problems under each assignment. In general, we observe that assignments with the input data types of ``string'' and ``int array'' have a lower mean of the error as compared to assignments with the input data type of ``int'' and ``boolean''. This difference is due to the former case involving more complex concepts and as a result, student-written codes exhibit similar kinds of misconceptions and bugs. Therefore, our approach yields a good set of generic test cases that are well-suited for these assignments. On the contrary, assignments involving simpler concepts like ``int'' and ``boolean'' result in student-written code that has a wide range of diverse bugs, which require a lot of test cases to capture them. Therefore, a single, limited set of $10-20$ test cases does not generalize well across all students and problems. We also note that the number of code submissions for categories ``int'' and ``boolean'' is double the number of code submissions for categories ``string'' and ``int array''. Our code pair selection strategy to select only \emph{three} code pairs to represent a particular assignment might not be effective for categories like ``int'' and ``boolean'' as they represent a wide variety of bugs owing to many student submissions. In the future, we can work on sampling diverse code pairs proportional to the number of code submissions for that assignment to ensure a lower mean error.

\paragraph{Analysis}
\figureref{fig:494-46} shows the error and frequency of code for problem 46 in assignment 494, ordered by their actual score. The problem is about checking if a given value is present in at least one of the adjacent pairs in a given array. Each blue point represents the mean error of the codes with a particular ground truth score. Each green bar represents the total number of codes corresponding to a particular range of the ground truth code score. The number of (almost) completely correct and incorrect codes, with the lowest and the highest scores, respectively, is much higher than the number of partially correct codes with intermediate scores. Most common student codes that are completely erroneous reflect programming conceptual misunderstanding including early returning of ``true'' or ``false'' from the method without processing the entire array or improper usage of control statements like ``continue'' and ``break''. The partially correct student codes that have intermediate scores are more subtle and do not generally portray misunderstanding in concepts. The errors are mostly about misunderstanding the problem and including incorrect array indices like ``i+2'' instead of ``i+1'' or incorrect conditions like ``=='' in the place of ``!=''. 

We observe that the generated set of test cases obtains the smallest error on codes with either the lowest or the highest score. This observation shows that most of the generated test cases either pass all the correct codes or fail on the worst-scored codes. Moreover, since the number of such codes (with either the highest or the lowest ground truth code score) is higher, the mean error turns out to be quite low. On the contrary, codes with intermediate ground truth scores are less frequent and have higher errors. This observation shows that the generated test cases still cannot always capture the subtle differences between partially correct codes with a few problems and completely correct code, which justifies our code pair selection approach that selects buggy codes with scores higher than the median and asks the LLM to generate test cases for them. 

\paragraph{Qualitative Example}
Listings \ref{lst:buggycode} and \ref{lst:correctcode} show a pair of buggy and correct codes, respectively, for problem 45 in assignment 502, and \tableref{tab:case-study} shows the problem statement and the generated test cases. The bug in the buggy code is that it only checks for 7 in the immediate next index after 6 and not anywhere after 6. As a result, the buggy code and the correct code have different outputs for test cases 2 and 3, while having the same output for test cases 1 and 4. This example shows that by explicitly referencing a buggy code, our approach for LLM prompting enables it to generate test cases that reflect student errors. 

\begin{figure}[hbtp]
\begin{minipage}{0.45\textwidth}
\label{lst:buggy}
\centering
\begin{lstlisting}[language=Java, caption={Buggy Code}, label=lst:buggycode]
public int sum67(int[] nums){
int sum = 0;
if (nums.length == 0){return 0;}
else{for (int i = 0; i < nums.length; i++){
if (nums[i] == 6){
if (nums[i + 1] == 7){i = i + 1;}}
else{sum = sum + nums[i];}}
return sum;}}
\end{lstlisting}
\end{minipage}%
\hfill
\begin{minipage}{0.5\textwidth}
\centering
\begin{lstlisting}[language=Java, caption={Correct Code}, label=lst:correctcode]
public int sum67(int[] nums){
int sum = 0; boolean sixMode = false;
for (int i = 0; i < nums.length; i++){
if(sixMode){
if (nums[i] == 7) sixMode = false;}
else if(nums[i] == 6) sixMode = true;
else sum += nums[i]; }
return sum;}
\end{lstlisting}
\end{minipage}
\end{figure}

\begin{table}[hbtp]
\small
\floatconts
  {tab:case-study}
  {\caption{A qualitative example that shows our method can generate test cases that reflect the bug in student-written buggy code. ``Buggy'' and ``Correct'' refer to the output of the buggy and the correct code respectively.}}
  {\scalebox{.87}{\begin{tabular}{p{10cm}lll}
  \toprule
  \bfseries Programming Question & \bfseries Test Case & \bfseries Buggy & \bfseries Correct\\
  \midrule
  \multirow{4}{10cm}{Given an int array, return the sum of the numbers in the array, except ignore sections of numbers starting with a 6 and extending to the next 7 (every 6 will be followed by at least one 7). Return 0 for no numbers.} 
  & $int[]\{6, 7, 1, 2, 3\}$ & 6 & 6 \\
  & $int[]\{6, 1, 2, 3, 4, 7\}$ & 17 & 0 \\
  & $int[]\{1, 2, 6, 3, 7\}$ & 13 & 3 \\
  & $int[]\{1, 2, 3, 4, 5, 6, 7\}$ & 15 & 15 \\
  \bottomrule
  \end{tabular}}}
\end{table}

\section{Conclusions and Future Work}

In this work, we propose a fully automatic iterative refinement-based approach for student code-guided test case generation using large language models. We craft a series of prompts that include selected representative student codes for automated test case generation and use compiler feedback to refine them. We evaluate our approach on a real-world student coding dataset and show that we can generate test cases that (sometimes) accurately capture student performance. There are plenty of avenues for future work. We can improve our approach by developing a diverse student code selection strategy that captures a wider variety of bugs, to further improve the diversity of the generated test cases. We also plan to conduct a human evaluation to assess the validity of our generated test cases and compare them to ones generated by computer science education experts. Along another direction, we can investigate whether test cases can be used as a formative assessment tool to measure students' programming knowledge. Then, we can use them in an adaptive testing setting, where the goal is to select and/or generate relevant test cases for students so as to assess their knowledge efficiently. Finally, we can also use these test cases to design modules that provide personalized feedback for student-written code to help them understand their errors and correct them.

\section{Acknowledgements}

The authors thank NSF for their support via grant DUE-2215193.

\appendix

\section{Iterative Prompt Engineering}

In our work, we propose a fully automatic method for generating test cases which involves prompting an LLM using feedback from a code compiler (Java compiler). To enable the process of automatic feedback, we had to ensure some engineering details. We ensure that the LLM prompt is in the form of a dictionary or a JSON-serializable object so that we can easily extract the test cases. Our feedback prompt is constructed to contain tabular information (represented as text) about each test case generated along with the output of the buggy and the correct code. This explicit way of representing feedback reduces hallucination in LLMs by giving point-wise test case feedback to the LLM. With regard to automatically executing the generated test cases against the buggy student codes, we observed that a few test cases triggered run-time errors including infinite loops thus causing the execution to never end. We use some engineering techniques like multi-threading and timing to cap the execution of a single test case on each student code. In case of run-time errors, we terminate the test case execution and provide feedback to the LLM regarding the run-time error so that it can rectify the test cases.

\bibliography{pmlr-sample}

\end{document}